\documentclass[runningheads]{llncs}

\usepackage{eccv}

\usepackage{eccvabbrv}

\usepackage{graphicx}
\usepackage{booktabs}
\usepackage{amsmath,amssymb}
\usepackage{multirow}
\usepackage[misc]{ifsym} 

\usepackage[accsupp]{axessibility}  
\usepackage{hyperref}

\usepackage{orcidlink}

\makeatletter
\renewcommand{\@fnsymbol}[1]{\ensuremath{\ifcase#1\or \dagger\or \ddagger\or
   \mathsection\or \mathparagraph\or \|\or **\or \dagger\dagger
   \or \ddagger\ddagger \else\@ctrerr\fi}}
\makeatother

\begin{document}

\title{SIGMA-Lane: Scale-pyramId Gated MAmba for Temporally Consistent Video Lane Detection} 

\titlerunning{SIGMA-Lane}

\author{Tiancheng Zhang\orcidlink{0009-0001-4746-1002} \and Yan Gao \and Xiangjie Kong\orcidlink{0000-0003-2698-3319} \and Guojiang Shen \and Jiaxin Du \and Mengmeng Wang\orcidlink{0000-0003-4577-6729}\thanks{Corresponding author:Mengmeng Wang.}}

\authorrunning{T. Zhang et al.}

\institute{Zhejiang Key Laboratory of Visual Information Intelligent Processing, College of Computer Science and Technology, Zhejiang University of Technology}

\maketitle

\begin{abstract}
Video lane detection requires predictions that remain stable across frames, yet severe vehicle occlusions can break temporal cues. In streaming recurrent models, corrupted observations may enter the hidden state and produce errors that persist into later frames. Existing occlusion-aware refinements usually provide obstacle masks as auxiliary inputs, so the state-update path is only indirectly protected. We propose SIGMA-Lane, which treats this failure mode as \emph{state contamination} in State Space Model (SSM)-based temporal modeling. SIGMA-Lane places occlusion-aware gates on the SSM write and residual-fusion paths, controlling how current observations enter temporal memory and are fused back after temporal propagation. After coordinate-consistent affine alignment, the model combines two complementary paths: SSM-consistent dual-gating for temporal filtering and Structural Spatial Retrieval (SSR) for recovering missing lane structure from aligned historical priors. Experiments on VIL-100 and OpenLane-V show improved temporal stability under heavy occlusion, with competitive F1 and mIoU scores.
\end{abstract}

\section{Introduction}

\begin{figure}[t]
  \centering
  \includegraphics[width=1\linewidth]{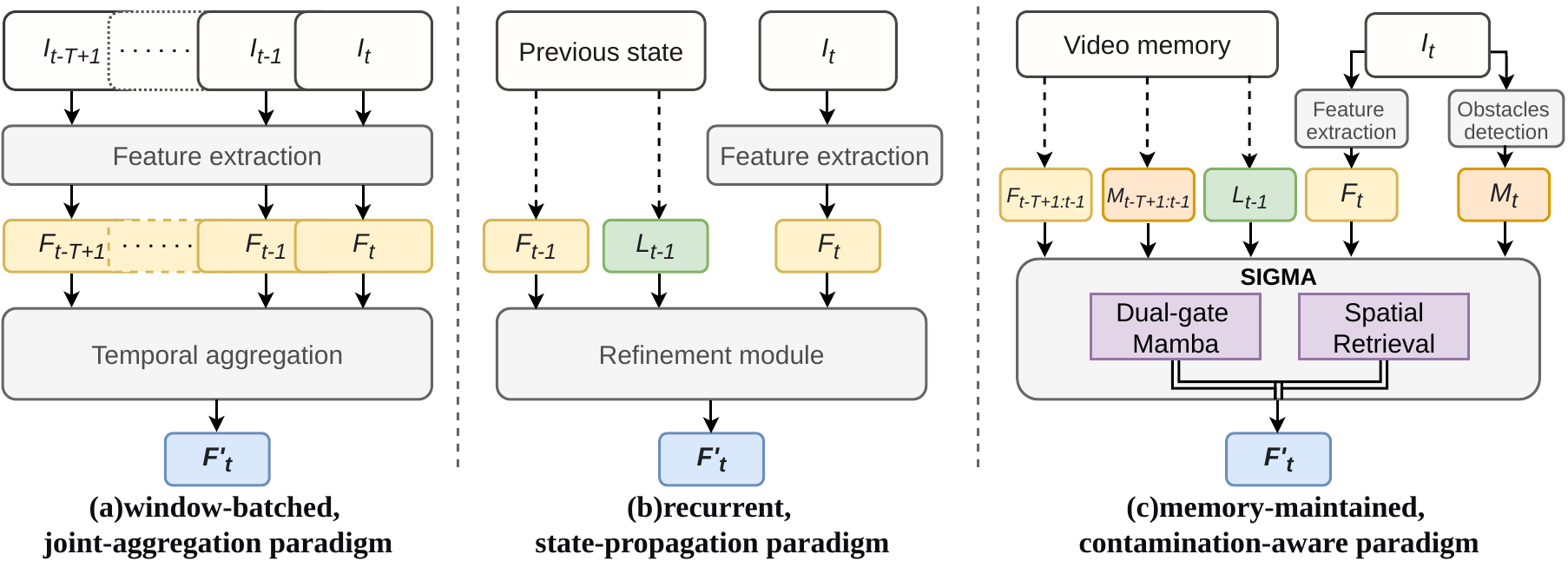}
  \caption{\textbf{Comparison of Video Lane Detection Paradigms.}
 The window-batched, joint-aggregation paradigm~(a) extracts features $F_{t-T+1:t}$ from a fixed temporal window and aggregates them jointly, incurring high latency without reusing past computation.
 The recurrent, state-propagation paradigm~(b) propagates the previous state $(F_{t-1}, L_{t-1})$ frame by frame, enabling streaming inference, but offers no mechanism to prevent corrupted observations from contaminating the hidden state under occlusion.
 Our memory-maintained, contamination-aware paradigm~(c) maintains a video memory bank of historical features $F$, obstacle masks $M$, and lane masks $L$. After lane-guided affine alignment, two paths process the aligned data: SSM-Consistent Dual-Gating for occlusion-aware temporal filtering and SSR for lane-guided spatial recovery.}
  \label{fig:approach}
\end{figure}

Lane detection provides structured road-boundary information for downstream planning and control in autonomous driving. Existing methods achieve strong per-frame accuracy with segmentation, parametric, or anchor-based representations~\cite{pan2018,zheng2022,tabelini2021CVPR}. However, in video, the detector must also keep the predictions consistent over time.
Two temporal designs are common. Window-batched methods~\cite{zhang2021,lanetca2025,mlmnet2023,wang2022video} jointly process a temporal window (Fig.~\ref{fig:approach}a). Recurrent methods~\cite{zou2020robust,zhang2021lane,jin2023,phnet2025} propagate a state frame by frame for streaming inference (Fig.~\ref{fig:approach}b). Both designs are vulnerable when large vehicles occlude lanes for several frames. If corrupted observations enter the temporal pipeline, they can affect later predictions unless the update itself suppresses them.

Existing recursive methods, such as OMR~\cite{jin2024omr}, feed obstacle masks into a ConvLSTM-based refinement module. This gives the model obstacle context, but it does not directly constrain how occluded features are written into the recurrent state. We examine this issue through the SSM update $h_t = \bar{A}_t h_{t-1} + \bar{B}_t x_t$, where the current observation enters the state through the write term. Mask-based noise filtering is common in vision, but in an SSM, this write term can still inject occlusion-corrupted features into the hidden state and leave residual errors across frames. Once stored, the corrupted feature becomes part of temporal memory and is reused by later predictions. We call this failure mode \emph{state contamination}.
This perspective motivates placing the obstacle prior on the SSM write path and the residual-fusion path, instead of using it only as an auxiliary input. It also exposes a second issue, \emph{coordinate inconsistency}: if features and masks are warped independently, the gate may no longer match the occluded region it is meant to modulate.

We propose SIGMA-Lane (Fig.~\ref{fig:approach}c), an SSM-based framework that gates the state update under occlusion. SSM-Consistent Dual-Gating reduces the effective write amplitude of $\bar{B}_t x_t$ at the input stage and controls residual fusion after temporal propagation. Lane-Guided Affine Warp applies the same affine transform to features, obstacle masks, and lane masks, keeping the gating signal in the same coordinate frame as the feature being modulated. SSR then recovers lane structure from aligned historical features with mask-guided cross-attention. A geometry-aware start token initializes the temporal state at the beginning of a stream. On VIL-100~\cite{zhang2021}, SIGMA-Lane achieves the best mIoU, F1@0.5, accuracy, and temporal stability scores, reducing $R_F/R_M$@0.5 to 0.023/0.030. On OpenLane-V~\cite{jin2023}, it obtains the best F1@0.5/F1@0.8 and the lowest persistent missing rate at the stricter threshold ($R_M$@0.8 = 0.384), while remaining competitive in mIoU (Tab.~\ref{table:vil100}, Tab.~\ref{table:openlane}).

Our contributions are:
\begin{itemize}
  \itemsep0mm
  \item We formulate state contamination as a failure mode in SSM-based video lane detection, and introduce a dual-gating mechanism that controls both the SSM write path and residual-fusion path under occlusion.
  \item We design a dual-path aggregation module with coordinate-consistent affine alignment, Mamba-based temporal propagation, and Structural Spatial Retrieval. The design separates temporal filtering from spatial recovery, addressing temporal-state contamination and loss of lane topology with different modules.
   \item On VIL-100 and OpenLane-V, SIGMA-Lane obtains competitive accuracy with lower flickering and missing rates. Ablations and controlled contamination analysis show that dual-gating accounts for a large share of the improvement.
\end{itemize}

\section{Related Work}

\noindent\textbf{Lane Detection Representations.}
Modern lane detectors use several representation strategies for complex road scenes. Segmentation-based methods~\cite{pan2018,qiu2022mfialane,hou2019road,hou2020inter,qin2020,zheng2021} treat lane detection as pixel-wise classification and improve spatial coherence with message passing~\cite{pan2018} or multi-scale aggregation~\cite{qiu2022mfialane,zheng2021}. Parametric methods~\cite{neven2018,liu2021,feng2022,wang2022,xu2022,xu2020curvelane,tabelini2021ICPR,lane2seq2024} model lanes as polynomials or B\'ezier curves; CondLaneNet~\cite{liu2021condlanenet}, for example, uses conditional convolution for instance-level prediction. Anchor-based detectors~\cite{tabelini2021CVPR,zheng2022,xiao2023,cheng2022dilane} use predefined line priors for localization. These single-frame methods perform well on static benchmarks, but they process each image independently and cannot use temporal context to infer lanes hidden by occlusion.

\noindent\textbf{Temporal Context in Video Lane Detection.}
Extending detection to videos reduces the instability of single-frame methods. Early recurrent designs~\cite{zou2020robust,zhang2021lane} use ConvLSTM or ConvGRU to model frame-to-frame dependencies. Recent transformers~\cite{vaswani2017}, including LaneTCA~\cite{lanetca2025} and MMA-Net~\cite{zhang2021}, use attention for global temporal context. Existing temporal aggregators still face a trade-off: recursive methods such as RVLD~\cite{jin2023} and OMR~\cite{jin2024omr} are efficient but can propagate corrupted observations, while attention-based models are more expensive. Current efficient solutions often use auxiliary motion correction~\cite{jin2023} or separate memory refinement~\cite{jin2024omr}, but they do not control how occluded evidence enters the temporal state. SIGMA-Lane targets this gap through a contamination-aware update rule.

\noindent\textbf{SSMs for Sequence Modeling.}
SSMs~\cite{gu2022s4,gu2023mamba} model long sequences with linear complexity. They have been used in generic vision tasks~\cite{zhu2024vision,vmamba2024,videomamba2024} and autonomous driving tasks, including 3D perception~\cite{mambabev2025}, motion planning~\cite{drivemamba2026}, and static lane detection~\cite{srlmamba2025}. Mamba adds selective scanning for content-aware sequence modeling. In occluded lane videos, however, contaminated observations may still enter the hidden state and accumulate over time. Our work complements Mamba's internal selectivity with an obstacle prior at the write and residual-fusion paths.

\section{Method}
\label{sec:method}

\subsection{Preliminary}
\label{ssec:prelim_framework}
We build on OMR's recursive video lane detection framework~\cite{jin2024omr}. We retain its encoder and decoder but replace its aggregation module. Given a current frame $I_t$, a ResNet18~\cite{he2016deep} encoder extracts a multi-scale feature map $F_t\in\mathbb{R}^{B\times C\times H\times W}$. The aggregation module refines $F_t$ into enhanced features $\tilde{F}_t$ using temporal-state propagation and spatial retrieval. Following the eigenlane paradigm~\cite{jin2022}, the decoder predicts a probability map $P_t$ and a coefficient map $C_t$. Lane instances are then extracted by NMS to obtain a binary lane mask $L_t$.

\begin{figure*}[!b]
  \centering
  \includegraphics[width=0.8\textwidth]{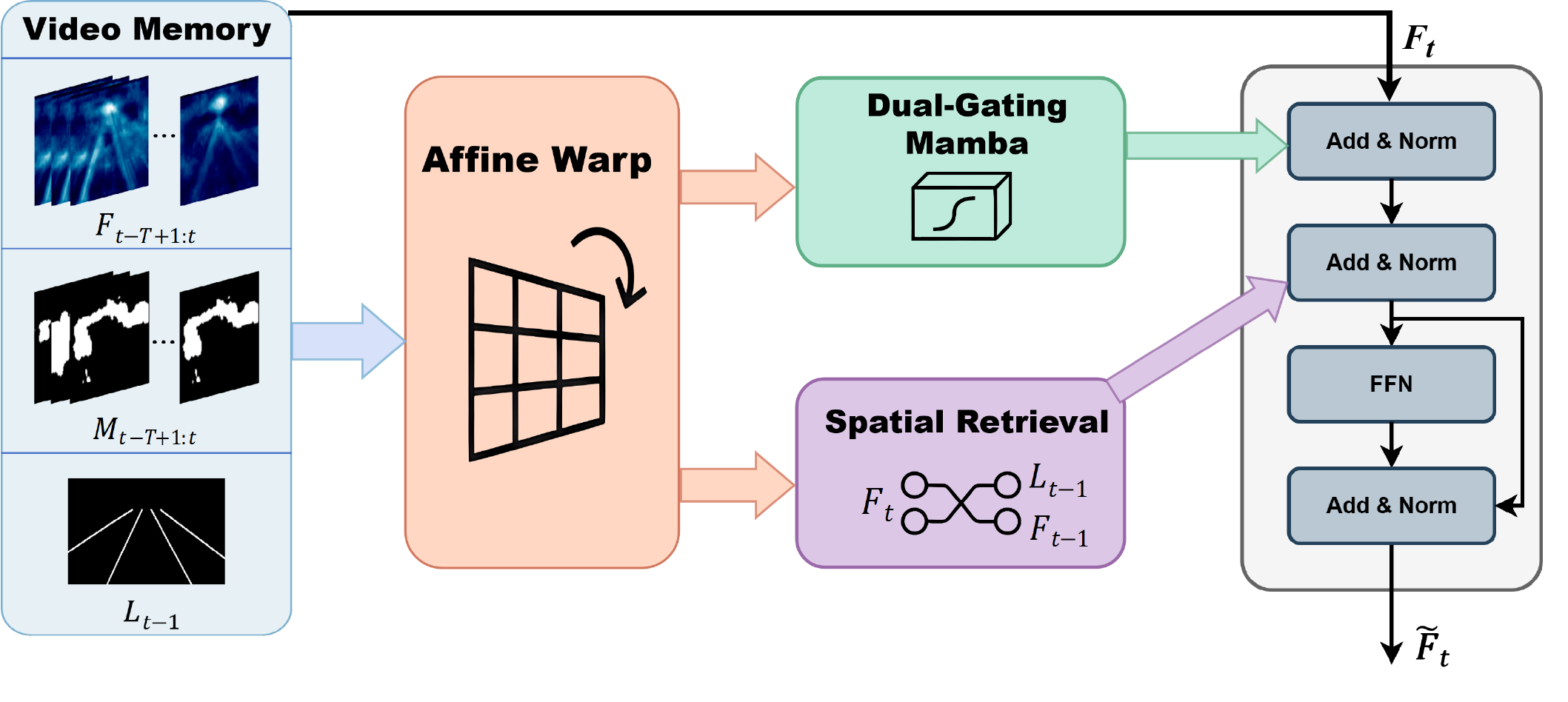}
  \caption{
 Overview of the SIGMA-Lane Aggregation Module. Video Memory stores historical features $F_{t-T+1:t}$, obstacle masks $M_{t-T+1:t}$, and the previous lane mask $L_{t-1}$. The Affine Warp module aligns historical data to current-frame coordinates. Warped features then pass through two paths: (1)~SSM-Consistent Dual-Gating with a scale-pyramid design for multi-scale temporal propagation and occlusion-aware gating, (2)~SSR where current features $F_t$ retrieve spatial context from warped $F_{t-1}$ enhanced by $L_{t-1}$ embeddings. The outputs are aggregated with current-frame features $F_t$ via Add\&Norm and FFN blocks, producing enhanced features $\tilde{F}_t$.
 }
  \label{fig:sigma_overview}
\end{figure*}

For streaming inference, a video memory bank stores the past $T$ frames' aggregated features and obstacle masks. With spatial positions flattened, these tensors are $\mathbf{F} \in \mathbb{R}^{(BHW) \times T \times C}$ and $\mathbf{M} \in \mathbb{R}^{(BHW) \times T \times 1}$. The bank also stores the previous lane mask $L_{t-1}$ for alignment and guidance. At the end of each frame, $\tilde{F}_t$, $M_t$, and $L_t$ are written back to memory. A geometry-aware start token initializes this recursive process at a cold start (Sect.~\ref{ssec:start_token}). OMR uses ConvLSTM for temporal refinement; we use Mamba2~\cite{dao2024mamba2} as the temporal update core and modify the aggregation module to address state contamination and coordinate inconsistency. SSM-Consistent Dual-Gating is the central temporal filter. Lane-guided alignment supplies coordinate-consistent masks, and SSR retrieves missing lane structure after gated temporal filtering. Fig.~\ref{fig:sigma_overview} shows the overall architecture.
\subsection{SSM-Consistent Dual-Gating}
\label{ssec:dual_gating}
\begin{figure}[t]
  \centering
  \includegraphics[width=0.90\linewidth]{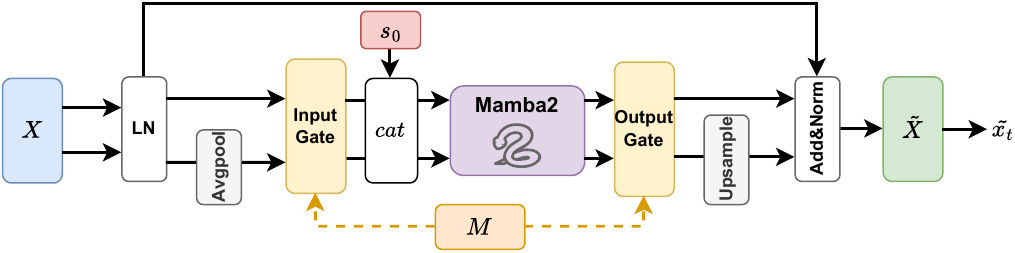}
  \caption{
 SSM-Consistent Dual-Gating with a scale-pyramid design. Input features $X$ pass through LN and are split into high- and low-resolution branches. Each branch applies Input Gate, concatenates the Geometry-Aware Start Token $s_0$, performs Mamba2~\cite{dao2024mamba2} temporal modeling, and applies Output Gate, with both gates modulated by the obstacle mask $M$. The low-resolution branch is upsampled and merged with the high-resolution branch through Add\&Norm to produce $\tilde{X}$. Finally, the current-frame features $\tilde{x}_t$ are extracted from $\tilde{X}$ and sent to subsequent modules.
 }
  \label{fig:dualgating}
\end{figure}

Observations from occluded regions are noisy. Directly writing them into the temporal state causes \emph{state contamination} that accumulates through recursive propagation. Starting from the linear update rule of SSMs,
\begin{equation}
 h_t = \bar{A}_t h_{t-1} + \bar{B}_t x_t,
    \label{eq:ssm_update}
\end{equation}
the central design question is where the obstacle prior enters the recursion. We gate two pathways where occlusion noise can enter the temporal output: the SSM write path, which controls what enters the temporal state, and the residual-fusion path, which controls how much current-frame evidence is mixed back after temporal propagation. The input gate acts before selective scanning and modulates the SSM write term. The output gate acts after temporal propagation and controls the residual path that mixes current features back into the temporal output. Mamba2's selective scan~\cite{dao2024mamba2} uses data-dependent parameters; conditioned on a given input sequence, each step can still be viewed as following the write-and-propagate structure in Eq.~\ref{eq:ssm_update}. This conditioned SSM view focuses the analysis on the write path affected by the external mask. Fig.~\ref{fig:dualgating} illustrates the module architecture.

\textbf{Input gating.}
We employ Mask-Aware Input Modulation:
\begin{equation}
 x'_t = x_t \odot (1 - m_t),
    \label{eq:maim}
\end{equation}
where $m_t\in[0,1]$ denotes the token-wise occlusion probability, broadcast over channels, and $\odot$ represents element-wise multiplication. The gate suppresses the input write term in Eq.~\ref{eq:ssm_update}:
\begin{equation}
 \bar{B}_t x_t \;\rightarrow\; (1-m_t)\,\bar{B}_t x_t.
\end{equation}
Highly occluded tokens therefore contribute little to the state write, while visible tokens are written normally.

This placement follows the contamination recursion. Let $x_t^{\star}$ denote the clean observation, $x_t=x_t^{\star}+\eta_t$ the occlusion-perturbed observation, and $\varepsilon_t$ the noise-induced state error. To isolate the explicit write path, we use a conditioned, linearized view in which the discretized parameters $\bar{A}_t$ and $\bar{B}_t$ are fixed around a trajectory. Without gating, the error evolves as
\begin{equation}
    \varepsilon_t = \bar{A}_t\varepsilon_{t-1} + \bar{B}_t\eta_t.
\end{equation}
With input gating, let $D_t$ denote the diagonal or channel-broadcast operator induced by $(1-m_t)$. Comparing the perturbed gated trajectory with its clean gated counterpart under the same local dynamics gives
\begin{equation}
    \varepsilon_t^{\mathrm{gate}} = \bar{A}_t\varepsilon_{t-1}^{\mathrm{gate}} + D_t\bar{B}_t\eta_t.
\end{equation}
When occlusion is severe ($m_t \to 1$), the entries of $D_t$ approach zero and the write-path noise term is suppressed before it enters the state.

For a contiguous occlusion segment $[s,s{+}L{-}1]$ and a later frame $t\ge s{+}L{-}1$, define the transition product
\begin{equation}
    \Phi(t,k)=\bar{A}_t\bar{A}_{t-1}\cdots\bar{A}_{k+1}, \qquad \Phi(t,t)=I .
\end{equation}
Unrolling the gated error recursion over the segment yields
\begin{equation}
    \varepsilon_t^{\mathrm{gate}}
    =
    \Phi(t,s{-}1)\varepsilon_{s-1}^{\mathrm{gate}}
    +
    \sum_{k=s}^{s+L-1}\Phi(t,k)D_k\bar{B}_k\eta_k .
    \label{eq:main_segment_unroll}
\end{equation}
Assume the conditioned dynamics are uniformly contractive, and the write matrices are bounded, with $\|\bar{A}_t\|\le\alpha<1$ and $\|\bar{B}_t\|\le\beta$. If $\|\eta_k\|\le\sigma$ on the segment and $\|D_k\|\le\gamma_k\le1$, Eq.~\ref{eq:main_segment_unroll} gives
\begin{equation}
    \|\varepsilon_t^{\mathrm{gate}}\|
    \le
    \alpha^{t-s+1}\|\varepsilon_{s-1}^{\mathrm{gate}}\|
    +
    \beta\sigma
    \sum_{k=s}^{s+L-1}\alpha^{t-k}\gamma_k .
    \label{eq:main_segment_gate_bound}
\end{equation}
With $\gamma=\max_{k\in[s,s+L-1]}\gamma_k$, this becomes
\begin{equation}
    \|\varepsilon_t^{\mathrm{gate}}\|
    \le
    \alpha^{t-s+1}\|\varepsilon_{s-1}^{\mathrm{gate}}\|
    +
    \beta\sigma\gamma\,
    \alpha^{t-s-L+1}\frac{1-\alpha^{L}}{1-\alpha}.
    \label{eq:main_segment_geo_bound}
\end{equation}
The ungated case is recovered by setting $D_k=I$ and $\gamma=1$. Under the conditioned, contractive write-path view above, a longer occlusion segment contributes a larger truncated geometric accumulation term, while input gating scales this term by the mask-dependent factor $\gamma$. The complete module, including the output gate, is evaluated empirically through ablation and controlled-contamination studies.

\textbf{Output gating.}
Input gating protects the state write, but the residual connection can still reintroduce noisy input features after temporal propagation. We control the residual fusion ratio at the output stage using the occlusion probability $M$ (a token-aligned obstacle mask, broadcast along channels). We summarize occlusion severity as a scalar $\bar{m}=\mathrm{GAP}(M)$ and compute the residual retention ratio
\begin{equation}
 \bar{m} = \mathrm{GAP}(M), \quad r = \sigma(\mathrm{MLP}(\bar{m})) \in (0,1),
\end{equation}
where $r$ is then broadcast to match the shape of $X_{\mathrm{in}}$. Let $X_{\mathrm{in}}$ denote the temporal block input and $X_{\mathrm{out}}$ the Mamba output:
\begin{equation}
 X_{\mathrm{base}} = X_{\mathrm{in}}+X_{\mathrm{out}},\qquad X_{\mathrm{hist}}=r\odot X_{\mathrm{in}}+X_{\mathrm{out}},
\end{equation}
\begin{equation}
 X_{\mathrm{time}}=(1-M)\odot X_{\mathrm{base}} + M\odot X_{\mathrm{hist}}.
    \label{eq:output_gate}
\end{equation}
The gate is initialized to a small value, close to 0.2, at the beginning of training for stability. Under occlusion, it down-weights the contaminated input $X_{\mathrm{in}}$ while keeping the propagated output $X_{\mathrm{out}}$. When $M=1$ (full occlusion), the output reduces to $r\odot X_{\mathrm{in}}+X_{\mathrm{out}}$; when $M=0$ (no occlusion), the standard residual $X_{\mathrm{in}}+X_{\mathrm{out}}$ is used.

\textbf{Relation to Mamba internal selectivity.} Mamba uses input-dependent parameters ($\Delta$, $B$, $C$) learned from data for content-aware selectivity. Our dual-gating adds an obstacle-mask prior at spatial positions likely to contain corrupted observations. The two mechanisms operate at different levels: Mamba's internal selectivity acts on content representations, while the external gates use the obstacle mask to modulate state writing and residual fusion at occluded positions.

\textbf{Multi-scale implementation.}
To aggregate information across the past $T$ frames, we flatten the feature sequence $\{x_{t-T+1},\ldots,x_t\}$ and occlusion sequence $\{m_{t-T+1},\ldots,m_t\}$ along spatial positions to obtain $X\in\mathbb{R}^{(BHW)\times T\times C}$ and $M\in\mathbb{R}^{(BHW)\times T\times 1}$. A parallel low-resolution branch captures broader context: features are downsampled by a factor $s$, processed through Mamba with dual-gating, and upsampled back. The outputs are fused via a learnable residual:
\begin{equation}
 \tilde{x}_t = x_{\mathrm{out}} + \alpha_{\mathrm{low}} \cdot \mathrm{Upsample}(\mathrm{Mamba}(\mathrm{Pool}_s(X))),
\end{equation}
where $\alpha_{\mathrm{low}}$ is initialized to a small value for stable training. The scale-pyramid branch combines fine local details with coarse context.

\subsection{Lane-Guided Affine Warp}
\label{ssec:lane_warp}
\begin{figure*}[t]
  \centering
  \includegraphics[width=\textwidth]{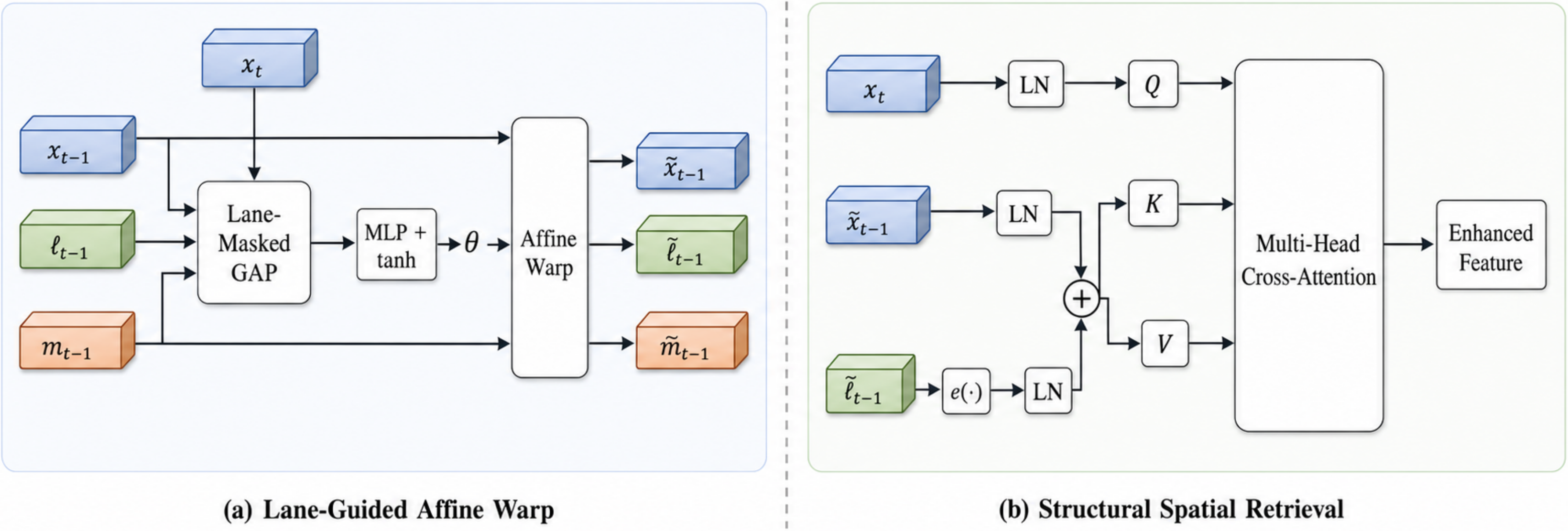}
  \caption{
 Lane-Guided Affine Warp and Structural Spatial Retrieval. (a)~In Lane-Guided Affine Warp, features $x_{t-1}$ and lane mask $\ell_{t-1}$ are fed into Lane-Masked GAP to compute lane-region-weighted statistics. Combined with current-frame features $x_t$, an MLP predicts affine parameters $p$, where $\tanh$ constrains the transformation magnitude. The Affine Warp operator synchronously transforms $x_{t-1}$, $\ell_{t-1}$, and $m_{t-1}$ using the same grid, yielding aligned outputs $\tilde{x}_{t-1}$, $\tilde{\ell}_{t-1}$, and $\tilde{m}_{t-1}$. (b)~In Structural Spatial Retrieval, current-frame features $x_t$ pass through LN to form the query $Q$. Warped features $\tilde{x}_{t-1}$ and lane-mask embedding $e(\tilde{\ell}_{t-1})$ are normalized via LN and summed to form key $K$ and value $V$. Multi-head cross-attention produces the enhanced feature output.
 }
  \label{fig:laneguided}
\end{figure*}

SSM-Consistent Dual-Gating applies temporal modeling after flattening spatial positions into token sequences. Each sequence therefore assumes that the same spatial index across frames refers to a consistent road location. Ego-motion and camera motion violate this assumption: a lane point or obstacle boundary may move to a different feature coordinate between adjacent frames. If such misaligned features are scanned along the $T$ dimension, the SSM can mix evidence from different physical locations. The obstacle gate can also shift away from the feature region it is meant to modulate. We therefore align the historical feature, lane prior, and obstacle mask before temporal propagation, using the synchronous strategy shown in Fig.~\ref{fig:laneguided}(a).

We use a local short-range affine transformation for lightweight 2D alignment between adjacent frames. This design matches the streaming state update, where information is propagated recursively and only the most recent transition must be synchronized before it enters the next temporal scan. In forward-facing driving videos, this low-dimensional parameterization is a compact approximation for synchronizing the feature, lane-prior, and obstacle-mask coordinates used by the gates. It is also less sensitive than dense optical flow~\cite{jin2023} to textureless or occluded regions. The resulting alignment supplies the token-aligned obstacle prior used by dual-gating. The lane mask $\ell_{t-1}$ provides an additional geometric prior: masked global average pooling (GAP) focuses transformation estimation on structurally stable lane areas while ignoring cluttered backgrounds.

Let $x_{t-1}$ and $x_t$ denote adjacent frame features. The affine parameters are predicted as:
\begin{equation}
 p=\tanh\!\left(\mathrm{MLP}\!\left([\mathrm{GAP}(x_{t-1};\ell_{t-1}),\ \mathrm{GAP}(x_t)]\right)\right),
    \label{eq:warp_param}
\end{equation}
where $[\cdot]$ denotes concatenation. We zero-initialize the final MLP layer so that the transformation starts from identity and avoids unstable early estimates. Reshaping $p$ yields a residual affine matrix $\Delta\Theta\in\mathbb{R}^{B\times 2\times 3}$. The final affine matrix is obtained by adding it to the identity affine $I_{2\times 3}$:
\begin{equation}
    \Delta\Theta = \mathrm{reshape}(p)\in\mathbb{R}^{B\times 2\times 3},\qquad
    \Theta = I_{2\times 3} + \Delta\Theta.
\end{equation}
The transformation synchronously warps features, lane masks, and obstacle masks via bilinear sampling:
\begin{equation}
 \tilde{x}_{t-1}=\mathrm{warp}(x_{t-1},\Theta),\quad
 \tilde{\ell}_{t-1}=\mathrm{warp}(\ell_{t-1},\Theta),\quad
 \tilde{m}_{t-1}=\mathrm{warp}(m_{t-1},\Theta).
    \label{eq:warp_all}
\end{equation}
This alignment keeps the occlusion signal $\tilde{m}_{t-1}$ and lane prior $\tilde{\ell}_{t-1}$ matched to the warped features $\tilde{x}_{t-1}$ and supports better aligned gating and structural guidance.

\subsection{Structural Spatial Retrieval}
\label{ssec:lane_xattn}
Dual-gating suppresses contaminated writes and residuals (Sect.~\ref{ssec:dual_gating}), but its temporal propagation can be insufficient when the historical state is weak and the current frame is heavily occluded. In this case, occluded regions still lack local lane evidence after gated temporal filtering. With aligned features $\tilde{x}_{t-1}$ and lane prior $\tilde{\ell}_{t-1}$ from Lane-Guided Affine Warp, SSR adds a spatial retrieval step to strengthen these missing structural cues. It uses cross-frame correspondence between the current frame and aligned history to recover structural evidence, as illustrated in Fig.~\ref{fig:laneguided}(b).

Current-frame features serve as the query, while warped previous-frame features form the key and value. To guide attention toward lane-relevant regions, we inject a learnable lane-mask embedding into the key/value features:
\begin{equation}
 Q=\mathrm{LN}(x_t),\qquad
 K=V=\mathrm{LN}(\tilde{x}_{t-1}) + \mathrm{LN}(e(\tilde{\ell}_{t-1})),
    \label{eq:lane_xattn_qkv}
\end{equation}
where $\mathrm{LN}$ denotes Layer Normalization and $e(\cdot)$ is a lightweight convolutional network mapping the single-channel mask $\tilde{\ell}_{t-1}$ to a $C$-dimensional feature space. Cross-attention follows the standard multi-head formulation~\cite{vaswani2017}:
\begin{equation}
 \mathrm{Attention}(Q,K,V)=\mathrm{softmax}\!\left(\frac{QK^{\top}}{\sqrt{d_k}}\right)V.
    \label{eq:xattn}
\end{equation}
The $QK^\top$ similarity matrix computes soft spatial correspondence, so each query position can retrieve structural cues from history without explicit offset prediction. Injecting the lane-mask embedding into the key/value features biases attention toward lane regions. Positions with stronger lane priors can receive higher matching scores, improving cross-frame completion under occlusion. SSR works with Dual-Gating (Sect.~\ref{ssec:dual_gating}), which handles recursive temporal state propagation.

\subsection{Geometry-Aware State Initialization}
\label{ssec:start_token}
\begin{figure}[t]
    \centering
    \begin{minipage}{0.48\linewidth}
        \centering
        \includegraphics[width=\linewidth]{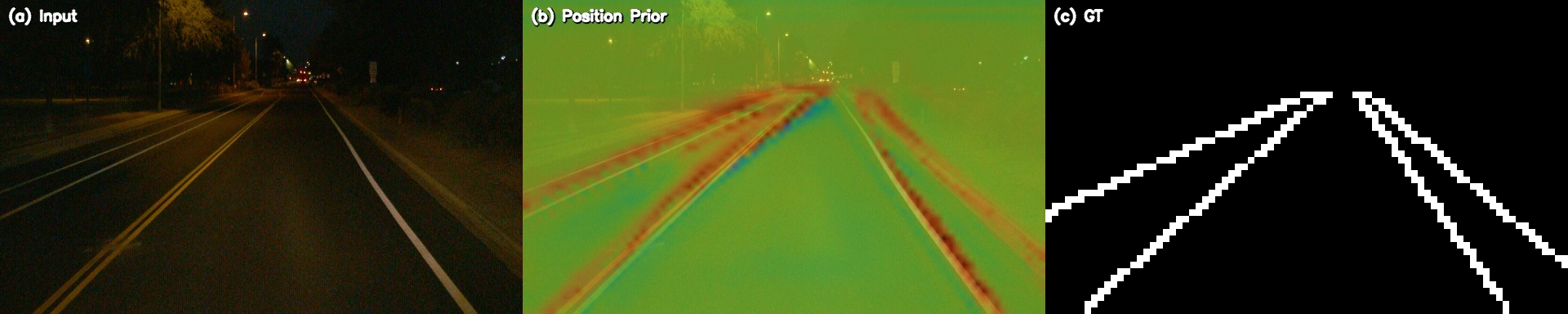}
    \end{minipage}
    \hfill
    \begin{minipage}{0.48\linewidth}
        \centering
        \includegraphics[width=\linewidth]{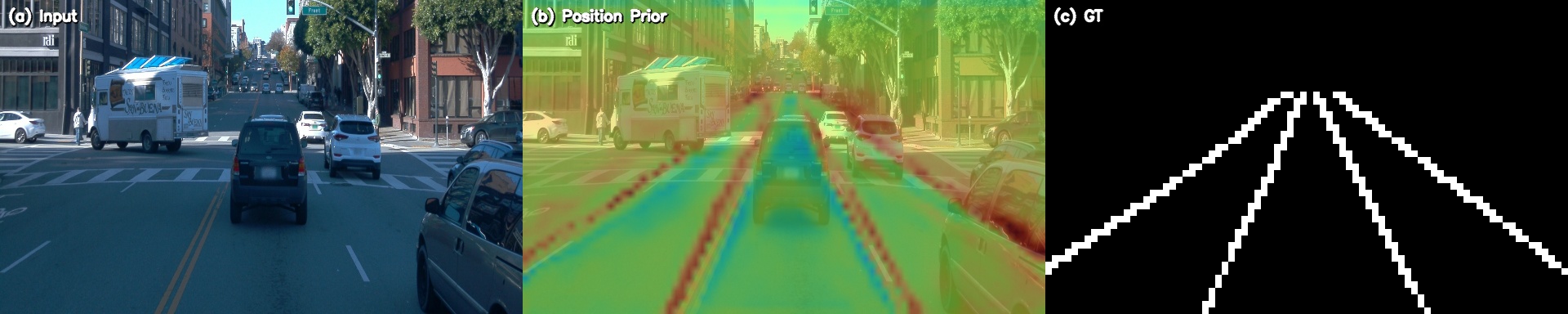}
    \end{minipage}
    \caption{Visualization of the spatial prior learned by Geometry-Aware State Initialization. Each strip shows the input image, position-prior heatmap, and ground-truth lanes. The learned prior is concentrated around lane-like structures and provides a cold-start bias before reliable temporal history is available.}
    \label{fig:start_token_vis}
\end{figure}

The previous modules control how observations enter an already running temporal state. At the beginning of a stream, however, this state is still uninitialized. Zero initialization ($h_0=0$) makes the first few state updates depend strongly on early frame observations, so an occluded or ambiguous first frame can dominate the initial memory. We use Geometry-Aware State Initialization to give the SSM a weak spatial prior before any image observation is written into the state.
We prepend a learnable start token $s$ to each spatial token sequence before Mamba processing. This token is read before the video frames and is not attenuated by the obstacle mask. To give it spatial semantics, we parameterize it as the sum of a context-agnostic shared token and a position-specific projection:
\begin{equation}
 s = \mathrm{token}_{\mathrm{shared}} + \mathrm{Proj}(\mathrm{PE}_{2D}),
\end{equation}
where $s\in\mathbb{R}^{(BHW)\times 1\times C}$ and $\mathrm{PE}_{2D}$ encodes normalized 2D coordinates. By processing $X^{+}=\mathrm{concat}(s, X)$, the SSM updates its hidden state $h_{-1} \to h_0$ using $s$ before seeing any video frames. This initialization complements dual-gating: the gates suppress contaminated writes, while the start token provides a geometry-aware state from which reliable temporal propagation can begin. Fig.~\ref{fig:start_token_vis} visualizes this learned prior, which tends to concentrate around lane-like regions.

\section{Experiments}
\label{sec:exp}

\subsection{Implementation Details}
The base detector described in Sect.~\ref{ssec:prelim_framework} is initialized from the pretrained OMR checkpoint and kept frozen. Only the temporal aggregation modules are trained. Following OMR, obstacle masks are produced by a frozen SegFormer-B5~\cite{xie2021segformer}; the segmenter is not updated during training. The temporal module uses Mamba2~\cite{dao2024mamba2} with an 8-frame memory window. It is trained with the same lane classification and coefficient regression losses as the base detector. Detailed hyperparameters, dataset descriptions, and metric definitions are provided in the supplementary material.

\subsection{Datasets and Evaluation Metrics}
We evaluate SIGMA-Lane on VIL-100~\cite{zhang2021} and OpenLane-V~\cite{jin2023}, two video lane benchmarks with persistent vehicle occlusion and diverse driving conditions. Following the CULane protocol~\cite{pan2018}, we report per-frame quality using mIoU, accuracy, and F1 at IoU thresholds 0.5 and 0.8. We also report temporal flickering and missing rates $R_F/R_M$ from~\cite{jin2023}, where lower values indicate fewer inconsistent detections across adjacent frames. The supplementary material gives the dataset details and metric definitions.

\begin{table}[!htb]\centering
    \renewcommand{\arraystretch}{0.80}
    \setlength{\tabcolsep}{2.6pt}
    \caption{Comparison on VIL-100. Best results are in \textbf{bold}, second best are \underline{underlined}.}
    \resizebox{\linewidth}{!}{
    \begin{tabular}[t]{l c c c c c c c}
    \toprule
 Method & Approach & mIoU$(\uparrow)$ & F1@0.5$(\uparrow)$ & F1@0.8$(\uparrow)$ & Accuracy$(\uparrow)$ & $R_F$@0.5$(\downarrow)$ & $R_M$@0.5$(\downarrow)$ \\
    \midrule
 LaneNet~\cite{neven2018}       & \multirow{4}{*}{Image-based} & 0.633 & 0.721 & 0.222 & 0.858 & -- & -- \\
 LSTR~\cite{liu2021}            &  & 0.573 & 0.703 & 0.131 & 0.884 & -- & -- \\
 LaneATT~\cite{tabelini2021CVPR}&  & 0.664 & 0.823 & -- & 0.912 & -- & -- \\
 DiLane~\cite{cheng2022dilane}    &  & 0.745 & 0.837 & 0.505 & 0.884 & -- & -- \\
    \midrule  
 MMA-Net~\cite{zhang2021}       & \multirow{6}{*}{Video-based} & 0.705 & 0.839 & 0.458 & 0.910 & 0.042 & 0.127 \\
 MLM-Net~\cite{mlmnet2023}      &  & 0.753 & 0.904 & \textbf{0.624} & -- & -- & -- \\
 RVLD~\cite{jin2023}            &  & 0.787 & 0.924 & 0.582 & -- & 0.038 & 0.050 \\
 PHNet~\cite{phnet2025}         &  & 0.783 & 0.915 & 0.615 & 0.908 & -- & -- \\
 OMR~\cite{jin2024omr}          &  & 0.774 & \underline{0.936} & 0.504 & 0.948 & \underline{0.026} & \underline{0.038} \\
 LaneTCA~\cite{lanetca2025}     &  & \underline{0.796} & 0.933 & \underline{0.621} & \underline{0.951} & 0.039 & 0.055 \\
    \midrule
 SIGMA-Lane             & Video-based & \textbf{0.801} & \textbf{0.940} & 0.595 & \textbf{0.956} & \textbf{0.023} & \textbf{0.030} \\
    \bottomrule
    \end{tabular}}
    \label{table:vil100}
\end{table}

\subsection{Comparative Results}
\sloppy

\begin{figure}[!b]
  \centering
  \includegraphics[width=0.92\linewidth]{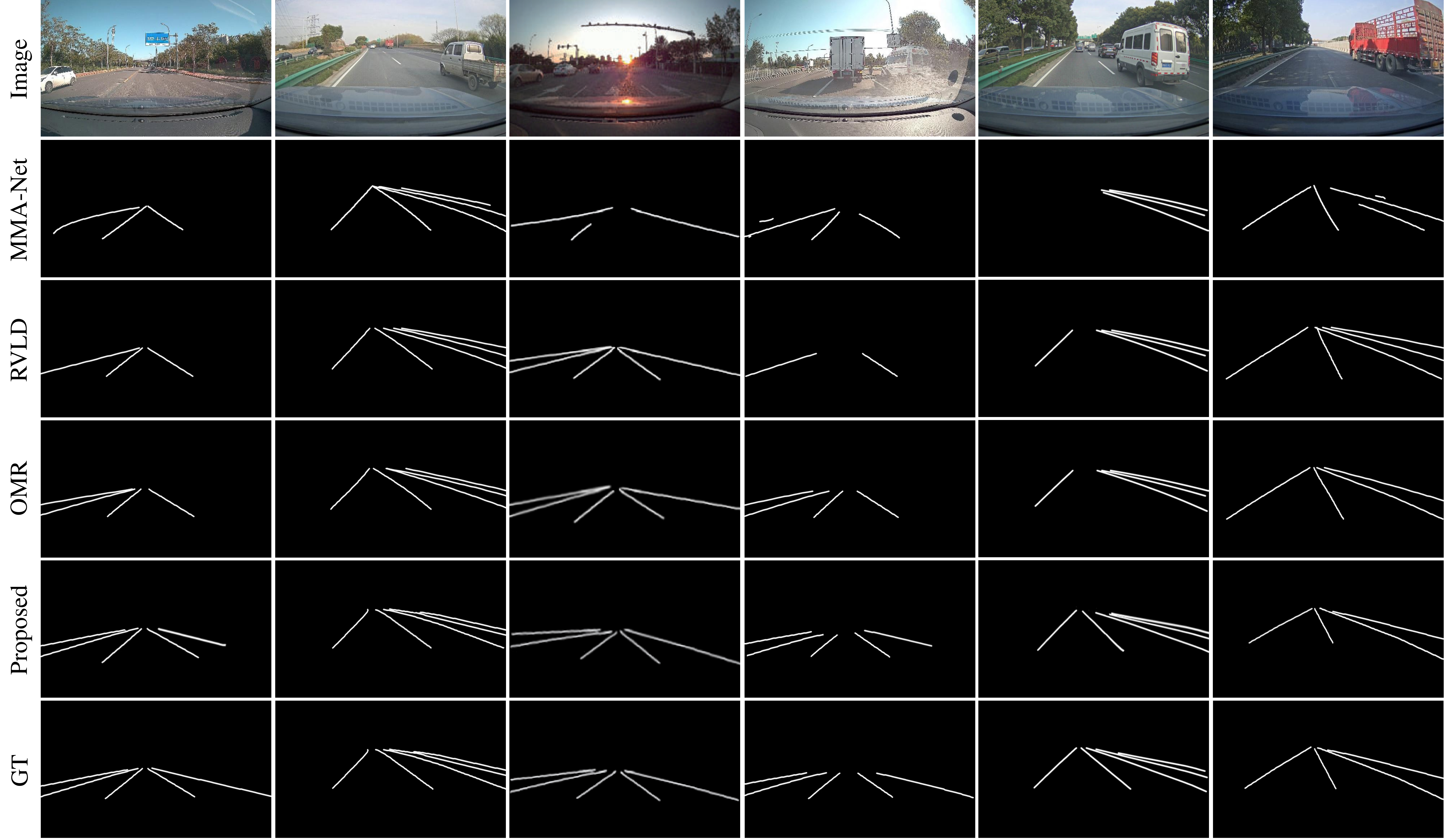}
  \caption{Comparison of lane detection results on the VIL-100 dataset.}
  \label{fig:qualitative_vil100}
\end{figure}

\noindent\textbf{VIL-100.}
Tab.~\ref{table:vil100} compares SIGMA-Lane with image-based~\cite{neven2018,liu2021,tabelini2021CVPR,cheng2022dilane} and video-based~\cite{zhang2021,mlmnet2023,jin2023,phnet2025,jin2024omr,lanetca2025} detectors. SIGMA-Lane obtains the best F1@0.5 (0.940), accuracy (0.956), and mIoU (0.801). Although SIGMA-Lane does not obtain the best F1@0.8, it has the lowest flickering and missing rates on VIL-100. Compared with LaneTCA, it reduces $R_F/R_M$ by 41.0\%/45.5\%. It also improves F1@0.8 by 0.091 over OMR, suggesting greater robustness when heavy occlusions suppress per-frame evidence. Fig.~\ref{fig:qualitative_vil100} shows examples where SIGMA-Lane preserves lane continuity when large vehicles obscure multiple lanes.

\begin{figure*}[!htb]
  \centering
  \includegraphics[width=0.92\linewidth]{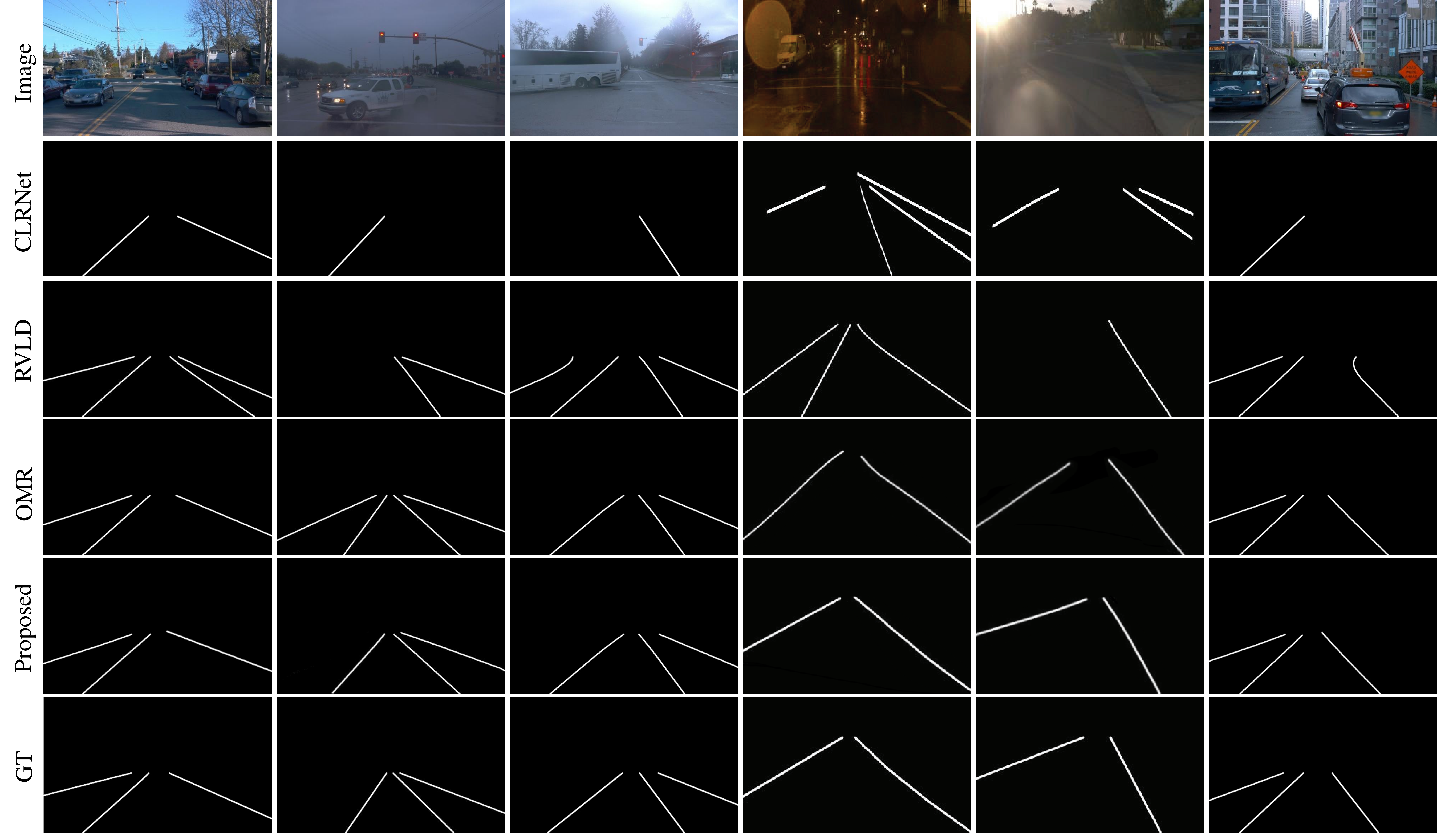}
  \caption{Comparison of lane detection results on the OpenLane-V dataset.}
  \label{fig:qualitative_openlane}
\end{figure*}

\begin{table*}[!htb]\centering
    \renewcommand{\arraystretch}{0.92}
    \caption{Comparison on OpenLane-V.}
    \resizebox{0.90\linewidth}{!}{
    \begin{tabular}[t]{l c c c c c c c c}
    \toprule
 Method & Approach & mIoU$(\uparrow)$ & F1@0.5$(\uparrow)$ & F1@0.8$(\uparrow)$ & ${\rm R}_{\rm F}$@0.5$(\downarrow)$ & ${\rm R}_{\rm M}$@0.5$(\downarrow)$ & ${\rm R}_{\rm F}$@0.8$(\downarrow)$ & ${\rm R}_{\rm M}$@0.8$(\downarrow)$ \\
    \midrule
 MFIALane~\cite{qiu2022mfialane}      & \multirow{5}{*}{Image-based} & 0.697 & 0.723 & 0.475 & 0.061 & 0.300 & 0.080 & 0.519 \\
 CondLaneNet~\cite{liu2021condlanenet}&  & 0.698 & 0.780 & 0.450 & 0.047 & 0.239 & 0.084 & 0.531 \\
 GANet~\cite{wang2022}                &  & 0.716 & 0.801 & 0.530 & 0.048 & 0.198 & 0.082 & 0.443 \\
 CLRNet~\cite{zheng2022}              &  & 0.735 & 0.789 & 0.554 & 0.054 & 0.224 & 0.086 & 0.430 \\
 DiLane~\cite{cheng2022dilane}          &  & 0.740 & 0.795 & 0.573 & 0.043 & 0.232 & 0.082 & 0.409 \\
    \midrule
 ConvLSTM~\cite{zou2020robust}        & \multirow{7}{*}{Video-based} & 0.529 & 0.641 & 0.353 & 0.058 & 0.282 & 0.091 & 0.574 \\
 ConvGRUs~\cite{zhang2021lane}        &  & 0.540 & 0.641 & 0.355 & 0.064 & 0.288 & 0.094 & 0.576 \\
 MMA-Net~\cite{zhang2021}             &  & 0.574 & 0.573 & 0.328 & 0.044 & 0.461 & 0.071 & 0.671 \\
 RVLD~\cite{jin2023}                  &  & 0.727 & 0.825 & 0.566 & \textbf{0.014} & \underline{0.167} & \textbf{0.051} & 0.406 \\
 OMR~\cite{jin2024omr}                &  & 0.742 & \underline{0.836} & 0.582 & \underline{0.016} & \textbf{0.162} & 0.055 & \underline{0.393} \\
 PHNet~\cite{phnet2025}               &  & 0.757 & 0.804 & \underline{0.588} & 0.025 & 0.196 & 0.064 & 0.399 \\
 LaneTCA~\cite{lanetca2025}           &  & \textbf{0.774} & 0.822 & 0.574 & 0.050 & 0.212 & 0.084 & 0.424 \\
    \midrule
 SIGMA-Lane                     & Video-based & \underline{0.761} & \textbf{0.838} & \textbf{0.591} & 0.018 & \textbf{0.162} & \underline{0.052} & \textbf{0.384} \\
    \bottomrule
    \end{tabular}}
    \label{table:openlane}
\end{table*}
\noindent\textbf{OpenLane-V.}
Tab.~\ref{table:openlane} reports results on the larger OpenLane-V benchmark. LaneTCA~\cite{lanetca2025} obtains the highest mIoU (0.774), while SIGMA-Lane obtains the highest F1@0.5/F1@0.8 (0.838/0.591) and the lowest persistent missing rate at the stricter threshold ($R_M$@0.8 = 0.384). This pattern reflects the different operating modes of the methods: LaneTCA aggregates a temporal window and achieves stronger average overlap, whereas SIGMA-Lane performs streaming state propagation and explicitly filters occluded writes. The gains at the stricter threshold match the intended use case: preserving valid lane instances under occlusion. Fig.~\ref{fig:qualitative_openlane} visualizes results across diverse conditions, and Fig.~\ref{fig:sigma_qual_analysis} traces one occluded case from intermediate maps to the recovered lane prediction. Additional examples of frame-to-frame behavior are provided in the supplementary material.

\begin{center}
\begin{minipage}{\linewidth}
    \centering
    \includegraphics[width=0.92\linewidth]{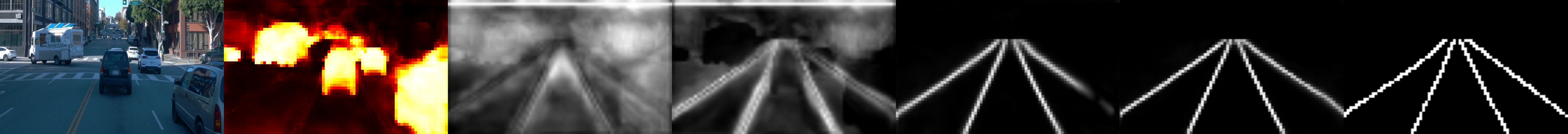}

    \makebox[0.92\linewidth]{%
        \small
        \makebox[0.12\linewidth][c]{$I_t$}\hfill%
        \makebox[0.12\linewidth][c]{$M_t$}\hfill%
        \makebox[0.12\linewidth][c]{$F_t$}\hfill%
        \makebox[0.12\linewidth][c]{$\tilde{F}_t$}\hfill%
        \makebox[0.12\linewidth][c]{$P_t$}\hfill%
        \makebox[0.12\linewidth][c]{$\tilde{P}_t$}\hfill%
        \makebox[0.12\linewidth][c]{GT}%
 }
  \captionsetup{hypcap=false}
  \captionof{figure}{
    Qualitative analysis of SIGMA-Lane under vehicle occlusion. Current-frame maps ($F_t, P_t$) show weakened lane responses in the occluded region, while refined maps ($\tilde{F}_t, \tilde{P}_t$) recover the lane topology and produce a prediction closer to GT. Maps are min-max normalized.
  }
  \label{fig:sigma_qual_analysis}
\end{minipage}
\end{center}

\subsection{Computational Cost}
Because SIGMA-Lane targets streaming video inference, we compare the computational cost of representative video lane detectors under the same $384{\times}640$ input setting in Tab.~\ref{table:cost}. All timing is measured on a single NVIDIA RTX 4090 in full-validation mode, using 50 warmup frames and all remaining frames for measurement. Model FPS measures the network-only forward pass, while E2E FPS includes post-processing and NMS inside the per-frame inference loop. Main Params exclude auxiliary ILD models and count only the temporal model used in the final stage. Agg. Params count only temporal aggregation modules. SIGMA-Lane has the lowest FLOPs and temporal aggregation parameters in this comparison, and runs faster than LaneTCA. OMR gives the highest throughput; SIGMA-Lane uses fewer FLOPs and aggregation parameters while obtaining higher accuracy.

\begin{table}[!htb]\centering
    \renewcommand{\arraystretch}{0.86}
    \setlength{\tabcolsep}{2.4pt}
    \caption{Computational cost comparison on OpenLane-V. Model FPS denotes network-only forward speed, while E2E FPS includes post-processing and NMS.}
    \resizebox{0.92\linewidth}{!}{
    \begin{tabular}{lccccc}
    \toprule
    Method & FLOPs(G)$(\downarrow)$ & Model FPS$(\uparrow)$ & E2E FPS$(\uparrow)$ & Main Params(M) & Agg. Params(M) \\
    \midrule
    OMR~\cite{jin2024omr}      & 54.6 & \textbf{215.96} & \textbf{141.15} & 7.15 & 3.54 \\
    LaneTCA~\cite{lanetca2025} & 28.9 & 41.57 & 28.48 & \textbf{1.62} & 0.96 \\
    SIGMA-Lane                 & \textbf{27.1} & 118.79 & 96.74 & 3.96 & \textbf{0.35} \\
    \bottomrule
    \end{tabular}}
    \label{table:cost}
\end{table}

\subsection{Ablation Studies}
We next isolate each component in the contamination-aware aggregation module. Row~1 in Tab.~\ref{table:ablation} replaces OMR's ConvLSTM with a standard Mamba2~\cite{dao2024mamba2} operator, with no gating or alignment. This gives a direct backbone-swap reference.

\begin{table}[!htb]\centering
  \footnotesize
  \setlength{\tabcolsep}{1.4pt}
  \renewcommand{\arraystretch}{0.80}
    \caption{Ablation studies on VIL-100. DG-I/O: Input/Output Gating, Warp: Affine Warp, SSR: Structural Spatial Retrieval, Token: Geometry-Aware Token.}
    
    \resizebox{\linewidth}{!}{
    \begin{tabular}[t]{c ccccc c c c c c}
    \toprule
    \# & DG-I & DG-O & Warp & SSR & Token & mIoU$(\uparrow)$ & F1@0.5$(\uparrow)$ & Accuracy$(\uparrow)$ & $R_F$@0.5$(\downarrow)$ & $R_M$@0.5$(\downarrow)$ \\
    \midrule
 1 &            &            &            &            &            & 0.775 & 0.926 & 0.942 & 0.052 & 0.084 \\
 2 & \checkmark &            &            &            &            & 0.781 & 0.930 & 0.947 & 0.036 & 0.045 \\
 3 &            & \checkmark &            &            &            & 0.778 & 0.929 & 0.946 & 0.040 & 0.053 \\
 4 & \checkmark & \checkmark &            &            &            & 0.787 & 0.935 & 0.950 & 0.024 & 0.032 \\
 5 & \checkmark & \checkmark & \checkmark &            &            & 0.794 & 0.937 & 0.952 & 0.025 & 0.035 \\
 6 & \checkmark & \checkmark & \checkmark & \checkmark &            & 0.798 & 0.938 & 0.953 & 0.024 & 0.033 \\
 7 & \checkmark & \checkmark & \checkmark & \checkmark & \checkmark & \textbf{0.801} & \textbf{0.940} & \textbf{0.956} & \textbf{0.023} & \textbf{0.030} \\
    \bottomrule
    \end{tabular}}
    \label{table:ablation}
\end{table}

\noindent\textbf{Dual-gating (rows 1--4).}
The baseline (row 1) is a competitive temporal backbone-swap reference. Input gating (row 2) attenuates the write term $(1{-}m_t)\bar{B}_t x_t$ to reduce corrupted state writes. Output gating (row 3) reduces the residual contribution of the noisy current frame at the fusion stage. Combining both gates (row 4) gives +0.012 mIoU (0.775$\to$0.787), which accounts for 46\% of the total mIoU gain (0.775$\to$0.801) in Tab.~\ref{table:ablation}. Dual-gating also reduces $R_F/R_M$ from 0.052/0.084 to 0.024/0.032. The full model reaches 0.023/0.030 after adding alignment, retrieval, and initialization.

\noindent\textbf{Lane-guided affine warp (row 5).}
Adding synchronous alignment gives +0.007 mIoU and keeps the obstacle mask $m_t$ aligned with the regions it should modulate. Without it, misalignment between warped features and the static mask can leak contaminated signals or over-suppress clean areas, which also hurts F1 and accuracy.

\noindent\textbf{Structural Spatial Retrieval (row 6).}
SSR adds +0.004 mIoU by retrieving lane-relevant spatial cues from warped historical features via lane-mask-guided cross-attention. It supplies structural information that gating alone cannot reconstruct, and also improves F1@0.5 and accuracy.

\noindent\textbf{Geometry-aware initialization (row 7).}
With dual-gating, alignment, and retrieval already enabled, the start token further improves the reliability of the initial temporal state. Row~7 gives the best values across all reported metrics, including the lowest missing rate, indicating that a geometry-aware prior helps recurrent aggregation start from lane-relevant spatial structure before sufficient history has accumulated.

\noindent\textbf{Ablation summary.}
These component ablations indicate a clear division of labor. Dual-gating provides the main temporal-stability gain by suppressing contaminated writes and residual fusion, while alignment, SSR, and the start token improve spatial recovery and initialization.

\subsection{Contamination and Occlusion Analysis}
We further analyze whether the improvement comes from the proposed contamination-aware use of obstacle priors, especially under corrupted memory and heavier occlusion.

\begin{table}[!htb]\centering
  \footnotesize
  \setlength{\tabcolsep}{3pt}
  \renewcommand{\arraystretch}{0.84}
  \caption{Mask and gate sensitivity on OpenLane-V validation. Deltas are measured relative to clean SIGMA-Lane. Lower is better for $\Delta R_F$ and $\Delta R_M$.}
  \resizebox{0.92\linewidth}{!}{
  \begin{tabular}{lcccc}
    \toprule
    Setting & $\Delta$F1$(\uparrow)$ & $\Delta$mIoU$(\uparrow)$ & $\Delta R_F(\downarrow)$ & $\Delta R_M(\downarrow)$ \\
    \midrule
    No DG-I/O; original $M$ & $-0.0180$ & $-0.0150$ & $+0.0362$ & $+0.0740$ \\
    Full SIGMA; dilate $M$ (7$\times$7) & $-0.0014$ & $-0.0010$ & $+0.0012$ & $+0.0024$ \\
    Full SIGMA; erase 20\% $M{=}1$ & $+0.0007$ & $-0.0005$ & $+0.0008$ & $-0.0010$ \\
    \bottomrule
  \end{tabular}}
  \label{table:mask_sensitivity}
\end{table}

\noindent\textbf{Mask and gate sensitivity.}
Tab.~\ref{table:mask_sensitivity} separates two possible explanations: whether the gain comes from having an obstacle prior, or from how the temporal model uses that prior. Perturbing the mask has a limited effect. A 7$\times$7 dilation changes F1/mIoU by only $-0.0014/-0.0010$, and erasing 20\% of obstacle pixels changes them by $+0.0007/-0.0005$. By contrast, removing both gates while keeping the original mask drops F1/mIoU by $-0.0180/-0.0150$ and increases $R_M$ by +0.0740. The result indicates that mask precision is not the main factor; the critical operation is placing the prior on the write and residual-fusion paths.

\noindent\textbf{Controlled contamination and occlusion severity.}
The supplementary material reports two additional analyses under stronger stress. Structured memory perturbations along occlusion boundaries increase $R_M$@0.5 from 0.162 to 0.182 for SIGMA-Lane, but from 0.236 to 0.367 for the ungated control, a 6.6$\times$ larger degradation without the gates. In the occlusion-stratified evaluation on OpenLane-V, SIGMA-Lane and OMR are comparable in Light frames, while SIGMA-Lane gains +0.028 mIoU over OMR in both Moderate and Heavy bins.

\section{Conclusion}
This paper studies video lane detection under persistent occlusion from the perspective of SSM state updates. We identify \emph{state contamination}: occlusion-corrupted observations can be written into temporal memory and affect later predictions. SIGMA-Lane addresses this failure mode by gating the SSM write and residual-fusion paths, with coordinate-consistent alignment, SSR, and geometry-aware initialization supporting temporal propagation and structural recovery. Experiments on VIL-100 and OpenLane-V show improved temporal stability with competitive per-frame detection quality. The ablations, mask sensitivity analysis, and controlled contamination tests indicate that the obstacle prior is most useful when it directly controls how information enters and leaves the temporal update. The current design operates in 2D image-feature space, where affine warping provides a local short-range alignment approximation. A natural extension is to combine contamination-aware state updates with 3D geometric cues, camera pose, or depth-aware motion for more complex driving scenes.

\subsubsection*{Acknowledgements}
This work was supported by the National Natural Science Foundation of China (Grant Nos. 62403429, 62476247, and 62402442), the Hangzhou Key Research and Development Program (Grant No. 2025SZDA0100), the Zhejiang Provincial Natural Science Foundation of China (Grant Nos. LQN25F030008 and LQ24F020038), and the Fundamental Research Funds for the Provincial Universities of Zhejiang (Grant No. G26081190006).

%
%
\bibliographystyle{splncs04}
\bibliography{references}

@String(CVPR  = {IEEE Conf. Comput. Vis. Pattern Recog.})

@String(ICCV  = {Int. Conf. Comput. Vis.})

@String(ECCV  = {Eur. Conf. Comput. Vis.})

@String(NeurIPS = {Adv. Neural Inform. Process. Syst.})

@String(ICML  = {Int. Conf. Mach. Learn.})

@String(ICLR  = {Int. Conf. Learn. Represent.})

@String(ACCV  = {Asian Conf. Comput. Vis.})

@String(AAAI  = {AAAI})

@String(ICPR  = {Int. Conf. Pattern Recog.})

@String(TCSVT = {IEEE Trans. Circuit Syst. Video Technol.})

@String(CVPR  = {CVPR})

@String(ICCV  = {ICCV})

@String(ECCV  = {ECCV})

@String(NeurIPS = {NeurIPS})

@String(ICML  = {ICML})

@String(ICLR  = {ICLR})

@String(ACCV  = {ACCV})

@String(ICPR  = {ICPR})

@String(TCSVT = {IEEE TCSVT})

@inproceedings{pan2018,
  title     = {Spatial as Deep: {S}patial {CNN} for traffic scene understanding},
  author    = {Pan, Xingang and Zhan, Xiaohang and Shi, Jianping and Luo, Ping and Wang, Xiaogang and Tang, Xiaoou},
  booktitle = AAAI,
  pages     = {7276--7283},
  year      = {2018}
}

@inproceedings{zheng2021,
  title     = {{RESA}: {R}ecurrent feature-shift aggregator for lane detection},
  author    = {Zheng, Tu and Fang, Hao and Zhang, Yi and Tang, Wenjian and Yang, Zheng and Liu, Haifeng and Cai, Deng},
  booktitle = AAAI,
  year      = {2021}
}

@inproceedings{qin2020,
  title     = {Ultra Fast Structure-aware Deep Lane Detection},
  author    = {Qin, Zequn and Wang, Huanyu and Li, Xi},
  booktitle = ECCV,
  year      = {2020}
}

@inproceedings{liu2021condlanenet,
  title     = {Cond{L}ane{N}et: {A} top-to-down lane detection framework based on conditional convolution},
  author    = {Liu, Lizhe and Chen, Xiaohao and Zhu, Siyu and Tan, Ping},
  booktitle = ICCV,
  year      = {2021}
}

@inproceedings{xu2020curvelane,
  title     = {Curve{L}ane-{NAS}: {U}nifying lane-sensitive architecture search and adaptive point blending},
  author    = {Xu, Hang and Wang, Shaoju and Cai, Xinyue and Zhang, Wei and Liang, Xiaodan and Li, Zhenguo},
  booktitle = ECCV,
  year      = {2020}
}

@inproceedings{tabelini2021ICPR,
  title     = {Poly{L}ane{N}et: {L}ane Estimation via Deep Polynomial Regression},
  author    = {Tabelini, Lucas and Berriel, Rodrigo and Paixao, Thiago M and Badue, Claudine and De Souza, Alberto F and Oliveira-Santos, Thiago},
  booktitle = ICPR,
  pages     = {6150--6156},
  year      = {2021}
}

@inproceedings{tabelini2021CVPR,
  title     = {Keep your eyes on the lane: Real-time attention-guided lane detection},
  author    = {Tabelini, Lucas and Berriel, Rodrigo and Paixao, Thiago M and Badue, Claudine and De Souza, Alberto F and Oliveira-Santos, Thiago},
  booktitle = CVPR,
  year      = {2021}
}

@inproceedings{zheng2022,
  title     = {{CLRN}et: {C}ross Layer Refinement Network for Lane Detection},
  author    = {Zheng, Tu and Huang, Yifei and Liu, Yang and Tang, Wenjian and Yang, Zheng and Cai, Deng and He, Xiaofei},
  booktitle = CVPR,
  pages     = {898--907},
  year      = {2022}
}

@inproceedings{liu2021,
  title     = {End-to-end lane shape prediction with transformers},
  author    = {Liu, Ruijin and Yuan, Zejian and Liu, Tie and Xiong, Zhiliang},
  booktitle = {WACV},
  year      = {2021}
}

@inproceedings{feng2022,
  title     = {Rethinking Efficient Lane Detection via Curve Modeling},
  author    = {Feng, Zhengyang and Guo, Shaohua and Tan, Xin and Xu, Ke and Wang, Min and Ma, Lizhuang},
  booktitle = CVPR,
  year      = {2022}
}

@inproceedings{cheng2022dilane,
  title     = {{DILane}: Dynamic Instance-Aware Network for Lane Detection},
  author    = {Cheng, Zhengyun and Zhang, Guanwen and Wang, Changhao and Zhou, Wei},
  booktitle = {ACCV},
  year      = {2022},
  pages     = {2075--2091}
}

@inproceedings{lane2seq2024,
  title     = {Lane2Seq: Towards Unified Lane Detection via Sequence Generation},
  author    = {Zhou, Kunyang},
  booktitle = CVPR,
  pages     = {16944--16953},
  year      = {2024}
}

@inproceedings{zhang2021,
  title     = {{VIL}-100: {A} New Dataset and A Baseline Model for Video Instance Lane Detection},
  author    = {Zhang, Yujun and Zhu, Lei and Feng, Wei and Fu, Huazhu and Wang, Mingqian and Li, Qingxia and Li, Cheng and Wang, Song},
  booktitle = ICCV,
  pages     = {15681--15690},
  year      = {2021}
}

@inproceedings{jin2023,
  title     = {Recursive Video Lane Detection},
  author    = {Jin, Dongkwon and Kim, Dahyun and Kim, Chang-Su},
  booktitle = ICCV,
  pages     = {8473--8482},
  year      = {2023}
}

@inproceedings{jin2024omr,
  title     = {{OMR}: Occlusion-Aware Memory-Based Refinement for Video Lane Detection},
  author    = {Jin, Dongkwon and Kim, Chang-Su},
  booktitle = ECCV,
  pages     = {129--145},
  year      = {2024}
}

@inproceedings{mlmnet2023,
  title     = {{MLM-Net}: Streamlined Multi-Lane Detection Network with Spatio-temporal Memory for Video Instance Lane Detection},
  author    = {Wang, Xiaoqin and Yin, Yunfei and Huang, Faliang and Bao, Xianjian},
  booktitle = {IEEE Int. Conf. Intell. Transp. Syst.},
  year      = {2023}
}

@article{lanetca2025,
  title   = {{LaneTCA}: Enhancing Video Lane Detection With Temporal Context Aggregation},
  author  = {Zhou, Keyi and Li, Li and Zhou, Wengang and Wang, Yonghui and Feng, Hao and Li, Houqiang},
  journal = TCSVT,
  volume  = {35},
  number  = {9},
  pages   = {8574--8585},
  year    = {2025}
}

@article{phnet2025,
  title   = {Parallel Heterogeneous Networks With Adaptive Routing for Online Video Lane Detection},
  author  = {Cheng, Zhengyun and Wang, Changhao and Zhang, Guanwen and Zhou, Wei},
  journal = {IEEE Trans. Intell. Transp. Syst.},
  volume  = {26},
  number  = {4},
  pages   = {5225--5235},
  year    = {2025}
}

@inproceedings{he2016deep,
  title     = {Deep residual learning for image recognition},
  author    = {He, Kaiming and Zhang, Xiangyu and Ren, Shaoqing and Sun, Jian},
  booktitle = CVPR,
  pages     = {770--778},
  year      = {2016}
}

@inproceedings{xie2021segformer,
  title     = {{SegFormer}: Simple and Efficient Design for Semantic Segmentation with Transformers},
  author    = {Xie, Enze and Wang, Wenhai and Yu, Zhiding and Anandkumar, Anima and Alvarez, Jose M and Luo, Ping},
  booktitle = NeurIPS,
  pages     = {12077--12090},
  year      = {2021}
}

@inproceedings{vaswani2017,
  title     = {Attention is all you need},
  author    = {Vaswani, Ashish and Shazeer, Noam and Parmar, Niki and Uszkoreit, Jakob and Jones, Llion and Gomez, Aidan N and Kaiser, {\L}ukasz and Polosukhin, Illia},
  booktitle = NeurIPS,
  year      = {2017}
}

@inproceedings{gu2022s4,
  title     = {Efficiently Modeling Long Sequences with Structured State Spaces},
  author    = {Gu, Albert and Goel, Karan and R{\'e}, Christopher},
  booktitle = ICLR,
  year      = {2022}
}

@article{gu2023mamba,
  title   = {Mamba: Linear-Time Sequence Modeling with Selective State Spaces},
  author  = {Gu, Albert and Dao, Tri},
  journal = {arXiv preprint arXiv:2312.00752},
  year    = {2023}
}

@inproceedings{dao2024mamba2,
  title     = {Transformers are {SSM}s: Generalized Models and Efficient Algorithms Through Structured State Space Duality},
  author    = {Dao, Tri and Gu, Albert},
  booktitle = ICML,
  pages     = {10041--10071},
  year      = {2024}
}

@article{videomamba2024,
  title   = {{VideoMamba}: State Space Model for Efficient Video Understanding},
  author  = {Li, Kunchang and Li, Xinhao and Wang, Yi and He, Yinan and Wang, Yali and Wang, Limin and Qiao, Yu},
  journal = {arXiv preprint arXiv:2403.06977},
  year    = {2024},
  note    = {ECCV 2024}
}

@article{vmamba2024,
  title   = {{VMamba}: Visual State Space Model},
  author  = {Liu, Yue and Tian, Yunjie and Zhao, Yuzhong and Yu, Hongtian and Xie, Lingxi and Wang, Yaowei and Ye, Qixiang and Jiao, Jianbin and Liu, Yunfan},
  journal = {arXiv preprint arXiv:2401.10166},
  year    = {2024},
  note    = {NeurIPS 2024}
}

@inproceedings{neven2018,
  title     = {Towards end-to-end lane detection: An instance segmentation approach},
  author    = {Neven, Davy and De Brabandere, Bert and Georgoulis, Stamatios and Proesmans, Marc and Van Gool, Luc},
  booktitle = {Intelligent Vehicles Symposium},
  year      = {2018}
}

@inproceedings{wang2022,
  title     = {A Keypoint-based Global Association Network for Lane Detection},
  author    = {Wang, Jinsheng and Ma, Yinchao and Huang, Shaofei and Hui, Tianrui and Wang, Fei and Qian, Chen and Zhang, Tianzhu},
  booktitle = CVPR,
  year      = {2022}
}

@inproceedings{xu2022,
  title     = {{RCL}ane: {R}elay Chain Prediction for Lane Detection},
  author    = {Xu, Shenghua and Cai, Xinyue and Zhao, Bin and Zhang, Li and Xu, Hang and Fu, Yanwei and Xue, Xiangyang},
  booktitle = ECCV,
  year      = {2022}
}

@inproceedings{jin2022,
  title     = {Eigenlanes: {D}ata-driven lane descriptors for structurally diverse lanes},
  author    = {Jin, Dongkwon and Park, Wonhui and Jeong, Seong-Gyun and Kwon, Heeyeon and Kim, Chang-Su},
  booktitle = CVPR,
  year      = {2022}
}

@inproceedings{xiao2023,
  title     = {{ADN}et: {L}ane Shape Prediction via Anchor Decomposition},
  author    = {Xiao, Lingyu and Li, Xiang and Yang, Sen and Yang, Wankou},
  booktitle = ICCV,
  year      = {2023}
}

@article{qiu2022mfialane,
  author  = {Qiu, Zengyu and Zhao, Jing and Sun, Shiliang},
  journal = {IEEE Transactions on Intelligent Transportation Systems},
  title   = {MFIALane: Multiscale Feature Information Aggregator Network for Lane Detection},
  year    = {2022},
  volume  = {23},
  number  = {12},
  pages   = {24263-24275}
}

@inproceedings{hou2019road,
  title     = {Learning Lightweight Lane Detection {CNN}s by Self Attention Distillation},
  author    = {Hou, Yuenan and Ma, Zheng and Liu, Chunxiao and Loy, Chen Change},
  booktitle = ICCV,
  year      = {2019}
}

@inproceedings{hou2020inter,
  title     = {Inter-Region Affinity Distillation for Road Marking Segmentation},
  author    = {Hou, Yuenan and Ma, Zheng and Liu, Chunxiao and Hui, Tak-Wai and Loy, Chen Change},
  booktitle = CVPR,
  year      = {2020}
}

@article{zou2020robust,
  title   = {Robust Lane Detection From Continuous Driving Scenes Using Deep Neural Networks},
  author  = {Zou, Qin and Jiang, Hanwen and Dai, Qiyu and Yue, Yuanhao and Chen, Long and Wang, Qian},
  journal = {IEEE Trans. Veh. Technol.},
  volume  = {69},
  number  = {1},
  pages   = {41--54},
  year    = {2020}
}

@article{zhang2021lane,
  title   = {Lane detection model based on spatio-temporal network with double convolutional gated recurrent units},
  author  = {Zhang, Jiyong and Deng, Tao and Yan, Fei and Liu, Wenbo},
  journal = {IEEE Trans. Intell. Transp. Syst.},
  volume  = {23},
  number  = {7},
  pages   = {6666--6678},
  year    = {2021}
}

@inproceedings{wang2022video,
  title     = {Video Instance Lane Detection via Deep Temporal and Geometry Consistency Constraints},
  author    = {Wang, Mingqian and Zhang, Yujun and Feng, Wei and Zhu, Lei and Wang, Song},
  booktitle = {ACM MM},
  year      = {2022}
}

@inproceedings{zhu2024vision,
  title     = {Vision Mamba: Efficient Visual Representation Learning with Bidirectional State Space Model},
  author    = {Zhu, Lianghui and Liao, Bencheng and Zhang, Qian and Wang, Xin and Liu, Wenyu and Wang, Xinggang},
  booktitle = ICML,
  year      = {2024}
}

@article{mambabev2025,
  title   = {{MambaBEV}: An Efficient {3D} Detection Model with {Mamba2}},
  author  = {You, Zihan and Wang, Ni and Wang, Hao and Zhao, Qichao and Wang, Jinxiang},
  journal = {arXiv preprint arXiv:2410.12673},
  year    = {2025}
}

@inproceedings{drivemamba2026,
  title     = {{DriveMamba}: Task-Centric Scalable State Space Model for Efficient End-to-End Autonomous Driving},
  author    = {Su, Haisheng and Wu, Wei and Song, Feixiang and Zhang, Junjie and Yang, Zhenjie and Yan, Junchi},
  booktitle = ICLR,
  year      = {2026}
}

@article{srlmamba2025,
  title   = {{SR-LMamba}: A Lane Detection Model for Complex Scenes Integrating Curvelet Transform with {Mamba} Architecture},
  author  = {Chen, Mingliang and Jia, Qinhao and Yang, Jing and Liu, Shuxian},
  journal = {PLOS ONE},
  volume  = {20},
  year    = {2025}
}
\end{document}